\documentclass{article} %
\usepackage{iclr2027_conference,times}

\usepackage{amsmath}
\usepackage{amssymb}
\usepackage{graphicx}
\usepackage{booktabs}
\usepackage{multirow}
\usepackage{array}
\usepackage{arydshln}
\usepackage{needspace}
\usepackage{algorithm}
\usepackage{algpseudocode}

\newcommand{\tl}[2]{\begin{tabular}[b]{@{}c@{}}#1\\#2\end{tabular}}
\newcommand{\yes}{$\checkmark$}
\newcommand{\no}{-}
\usepackage{capt-of}
\usepackage{hyperref}
\usepackage{url}
\usepackage{etoolbox}
\makeatletter
\patchcmd{\@maketitle}{\lhead{Published as a conference paper at ICLR 2027}}{\lhead{Preprint. Under review.}}{}{\PackageError{main}{header patch (final) failed}{}}
\patchcmd{\@maketitle}{\lhead{Under review as a conference paper at ICLR 2027}}{\lhead{Preprint. Under review.}}{}{\PackageError{main}{header patch (review) failed}{}}
\makeatother
\newcommand{\projecturl}{https://su-terry.github.io/cepheid-website/}

\graphicspath{{figs/}}

\title{Cepheid: Decoupling Scale from Structure\\ for Cross-Family Motion Completion}

\author{Chih-Hai Su$^{1}$ \quad Tsung-Wei Ke$^{2}$ \quad Yu-Lun Liu$^{1}$\\
$^{1}$National Yang Ming Chiao Tung University\\
$^{2}$Department of Computer Science and Information Engineering, National Taiwan University}

\iclrfinalcopy
\begin{document}

\maketitle

\begingroup
\centering
\vspace{-18pt}
\includegraphics[width=\linewidth]{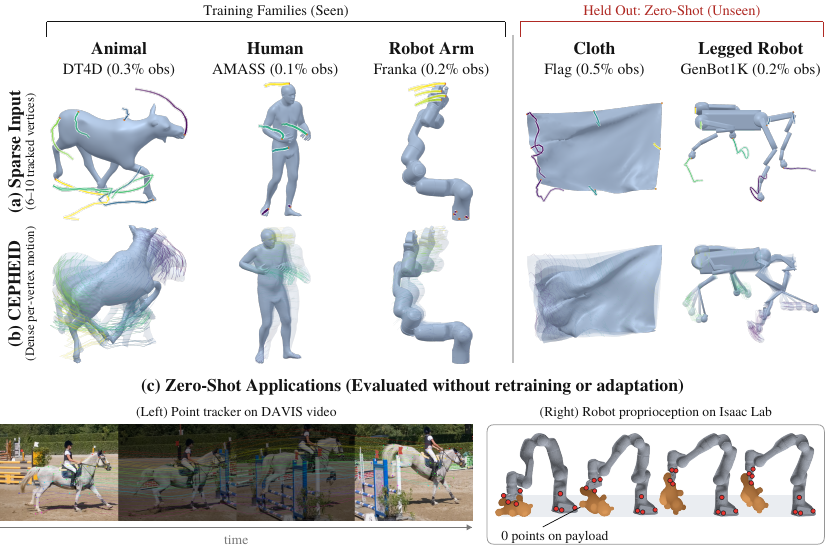}\par
\vspace{-10pt}
\captionof{figure}{\textbf{Our unified model, \textsc{Cepheid}, completes dense motion on various geometry families, including those never seen in training.}
(a)~Input: a mesh and the trajectories of 6--10 of its vertices (0.09--0.5\% of the mesh); no skeleton, registration, or cross-mesh correspondence is given.
(b)~Output: the motion \textsc{Cepheid} predicts for every vertex, drawn as per-vertex trajectories with the predicted poses ghosted. Every panel in (b) is a prediction, not ground truth.
The model is trained on the three families on the left only and is evaluated zero-shot on the cloth and the legged robot.
(c)~The same checkpoint, without retraining, densifies a point tracker's output (CoTracker3) on a DAVIS video (left) and completes the motion of a soft body lifted by a Franka arm from eight proprioceptive points on the robot and none on the payload (right).}
\label{fig:embodiments}\par
\vspace{2pt}
\endgroup

\raggedbottom
\begin{abstract}
Sparse-to-dense motion completion recovers dense mesh motion from sparse vertex displacements. Residual priors trained on one geometry family frequently fail to beat parameter-free harmonic extension off-family. Moreover, cross-family training meets a scaling barrier: residual amplitude varies by more than an order of magnitude across geometry families and collapses as observations become denser. Since this scale depends on unknown target motion, it cannot be measured directly at test time. We decouple residual structure from physical scale by learning a dimensionless prior, estimating amplitude from observables and a lightweight correction fitted only on training sources. Our unified model surpasses harmonic extension across all evaluated families, including unseen cloth and legged bodies, without skeletons, registration, or cross-mesh correspondence. The same prior applies without retraining to point-tracked animal video and a robot grasping a soft body.

Project page: \expandafter\url\expandafter{\projecturl}
\end{abstract}
\flushbottom

\vspace{-1mm}
\section{Introduction}

Without skeletons, registration, or cross-mesh correspondence, sparse-to-dense motion completion infers dense mesh motion from sparse displacements under partial visibility. Mesh connectivity yields a closed-form discrete harmonic extension, solving the graph Dirichlet problem in one sparse linear solve with observed vertices as boundary conditions. We use harmonic extension as the base for the learned residual and as the parameter-free control. Predictions worse than this control reveal that completion would have been more accurate without learning.

Priors trained on one geometry family fail to transfer reliably. Across three single-family models (DT4D animals, AMASS humans, Franka arm) and a unified model, seven of twelve single-family transfers fail to improve on harmonic extension. On cloth, the AMASS and Franka priors are 25.5\% and 7.8\% worse than the base, respectively. Because all share a frozen amplitude rule, these are failures of the residual priors. Training jointly across families also encounters extent-normalized residual amplitudes varying by $18\text{--}39\times$ across the three training families at matched density and falling by more than an order of magnitude across densities. Lacking target motion at inference, the residual cannot be normalized by its own scale.

In \textsc{Cepheid}, we decouple residual structure from physical scale, learning a dimensionless residual motion prior conditioned on geometry and harmonic extension while recovering amplitude without requiring target motion (Figure~\ref{fig:embodiments}). On cloth, completion error with recovered amplitude is within 1--8\% of a target-informed oracle (the per-condition optimal scalar that requires ground truth). Our unified model stays below harmonic extension on all five evaluated families, cutting error by 15.6\% on an unseen cloth sheet and 10.1\% on unseen procedural legged bodies.

Our contributions are threefold:
\begin{enumerate}
    \item \textbf{Harmonic extension as a general base and control.} We formulate motion completion around a Dirichlet solve on the mesh graph with the observed vertices as boundary conditions, providing the base for the residual and the control that seven of the twelve single-family transfers fail to improve on.
    \item \textbf{Target-independent amplitude recovery.} We recover physical amplitude without target motion, bringing completion error on cloth within 1--8\% of the error with the target-informed oracle across observation densities.
    \item \textbf{A shared dimensionless residual prior.} We learn a dimensionless residual prior with flow matching \citep{lipman2023flow, liu2023rectifiedflow, albergo2023interpolants}, conditioned on geometry and the harmonic extension. We are not aware of prior work showing that a single learned prior can complete dense per-vertex motion from sparse, unlabeled vertex displacements across geometry families, without skeletons, templates, or cross-mesh correspondence, and transfer zero-shot to families absent from training. \textsc{Cepheid} introduces such a prior. Fine-tuning to a fixed three-marker gripper placement more than halves the error of the same checkpoint evaluated zero-shot (Appendix~\ref{app:finetune}). The same checkpoint applies without retraining to point-tracked animal video and to a robot grasping a soft body.
\end{enumerate}

\vspace{-1mm}
\section{Related Work}
\label{sec:related_work}

\vspace{-1mm}
\paragraph{Deformation from sparse handles.}
Harmonic extension solves the graph-Laplacian Dirichlet problem with observed vertices as boundary conditions, as in label propagation \citep{zhu2003semi}. Handle-based deformation minimizes quadratic energies \citep{botsch2008linear}: Laplacian editing \citep{LaplacianMeshEditing:2004}, biharmonic deformation and weights \citep{botsch2004intuitive, jacobson2011bounded}, as-rigid-as-possible energies \citep{ARAP_modeling:2007, baieri2024implicit}, and embedded deformation graphs \citep{sumner2007embedded}; deformation transfer needs cross-mesh correspondence \citep{sumner2004deformation}. Learned handles predict cages \citep{yifan2020neural}, keypoints \citep{jakab2021keypointdeformer}, meta-handles \citep{liu2021deepmetahandles}, per-face Jacobians \citep{aigerman2022njf}, or local detail \citep{sundararaman2024deformation, maggioli2024sshade}, or follow 2D diffusion priors \citep{yoo2024plausible}, each trained per category. NSDP \citep{tang2022neural} regresses dense motion from sparse handles per class with per-frame scale normalization; we retrain it on our split (\S\ref{sec:baselines}). Per-scene optimization needs no training set \citep{DOMA, song2025spline, wang2025canfields, bozic2021neural}; DOMA \citep{DOMA} is evaluated in \S\ref{sec:baselines} and Appendix~\ref{app:connectivity_free}. Other sparse-input methods need facial landmarks on a template \citep{otberdout2022sparse}, labeled keypoints \citep{muralikrishnan2025smf}, skeletons \citep{muralikrishnan2024trj}, partial views, depth, or point clouds \citep{2024arXiv241208511M, maheshwari2023transfer4d, newcombe2015dynamicfusion, bozic2020deepdeform, palafox2021npms, li2022non, han2018robust}, or images \citep{collins2023ca2tnetcategoryagnostic3darticulation}, or reconstruct 4D shapes without meshes \citep{cao2024motion2vecsets, niemeyer2019occupancy, lei2022cadex}. We instead take unlabeled displacements on a mesh, use harmonic extension as a parameter-free base and control, and learn one residual prior across families.

\vspace{-1mm}
\paragraph{Family-specific and unified motion models.}
Deformable families usually come with their own representation: body models for humans and quadrupeds \citep{SMPL:2015, Zuffi:CVPR:2017}, the Unified Robot Description Format for robots, learned skeletal articulation for characters \citep{li2021learning}, or a cloth or soft-body simulator \citep{narain2012adaptive, lai2026diffphd}. Inside one family, sparse-to-dense recovery is mature: full-body human motion is recovered from three trackers \citep{dittadi2021full, aliakbarian2022flag, du2023agrol, castillo2023bodiffusion, feng2024stratified, tang2024unified}, optical markers \citep{ghorbani2021soma}, or sparse 4D input \citep{marsot2022structured, marsot2022representing}, but only through the body model. Unified models span categories with one network: UniMate \citep{mou2026unimate}, AnyTop \citep{gat2025anytop}, and OmniMotionGPT \citep{yang2024omnimotiongpt} need an articulated motion chain, ScanMove \citep{BESNIER2025104409} a driving motion, MeshLoom \citep{chen2026meshloom} a complete mesh per frame, and AnimateAnyMesh \citep{wu2025animateanymesh, wu2026animateanymesh++} a text prompt. Skeleton-free transfer \citep{wang2020neural, liao2022skeletonfree, wang2023zeroshotpose, wang2023hmc, zhang2024tapmo, chen2025motion2motion, zhang2026semantic} needs a source pose, motion, or bone correspondence; automatic rigging \citep{zhang2025unirig, song2026puppeteer, chen2026vips} and trajectory control \citep{liu2026follow} reintroduce a skeleton; video- or text-driven animation \citep{jiang2024animate3d, uzolas2025motiondreamer, shi2025drive, sabathier2026actionmesh} conditions on a prompt or a video. Closest to us, CANOR \citep{he2025category} drives a mesh from learned blobs but trains per category, and CompletionAny4D \citep{nishizawa2026feed} in-betweens temporally sparse but spatially complete keyframes, leaving partial regions and point trajectories to future work. Cross-embodiment robot policies \citep{gupta2022metamorph, bohlinger2024urma, doshi2024crossformer, sferrazza2024body} motivate our held-out legged family. Our only signal is sparse motion observations on the target.

\vspace{-1mm}
\paragraph{Learning one prior across families.}
Source diversity can help transfer \citep{fang2022clipdata}, but distribution mismatch causes negative transfer \citep{albuquerque2021generalizingunseendomainsdistribution, 8953565, wang2023dgsurvey}; a mesh tracker trained on six objects underperforms per-object training \citep{mansour2023fast}. These findings assume shared input spaces, whereas our families differ in topology, vertex count, and correspondence, so we test diversity empirically. Joint training also needs scale normalization by a training-set statistic \citep{lam2023learning} or an instance-level observable \citep{kim2022reversible, passalis2019deep, SALINAS20201181, yin2023metric3d, liu2022nonstationary, mccabe2024mpp, zuo2025omni}. \citet{3648699.3648808} write shared forms as a gauge length times a dimensionless function \citep{buckingham1914physically}, and dimensionless groups can be discovered from data \citep{bakarji2022dimensionally, xie2022data}. We apply this to the residual after harmonic extension, in the spirit of residual physics \citep{zeng2020tossingbot}, and recover the unit from observed displacement dispersion plus a dimensionless correction (\S\ref{sec:amplitude_recovery}, \S\ref{sec:amplitude_eval}). Our backbone follows graph-network simulators \citep{sanchez2020learning}, whose multiscale \citep{fortunato2022multiscale, cao2023efficient} and attention-augmented variants \citep{janny2023eagle, hao2023gnot, wu2024transolver} carry information between distant observed vertices. Unlike flow matching on manifolds \citep{chen2024flow}, latent-point mesh diffusion \citep{lyu2023controllable}, or native 4D generation \citep{li2025ss4d}, we sample a dimensionless residual field in Euclidean space with the standard flow-matching objective.

\vspace{-1mm}
\paragraph{Point tracking and 4D reconstruction.}
Point trackers \citep{harley2022pips, doersch2023tapir, karaev2024cotracker, wang2023omnimotion, lemoing2024dot, xiao2024spatialtracker, koppula2024tapvid3d, xiao2025spatialtrackerv2, zheng2023pointodyssey, cho2024locotrack} follow points identifiable in a query frame \citep{3600270.3601259}; dense 3D trackers \citep{ngo2025delta, ngo2025deltav2, nam2026trackcraft3r} and monocular 4D reconstruction \citep{liu2023robust, wang2025shapeofmotion, fan2025spectromotion, shih2025prior, chen2025v2m4, chan2026adagar, lin2026sm4rt} lift every visible pixel but start from a video. We complete every mesh vertex, including those never seen by the camera, from any observation source; tracking is an input rather than a baseline, and \S\ref{sec:applications} uses CoTracker3 \citep{karaev2024cotracker3}.

\vspace{-1mm}
\section{Method}

\vspace{-1mm}
\subsection{Problem Setting and the Harmonic Residual}
\label{sec:problem_setting}
\vspace{-1mm}

\begin{figure}[t]
\centering
\vspace{-3mm}
\includegraphics[width=\textwidth]{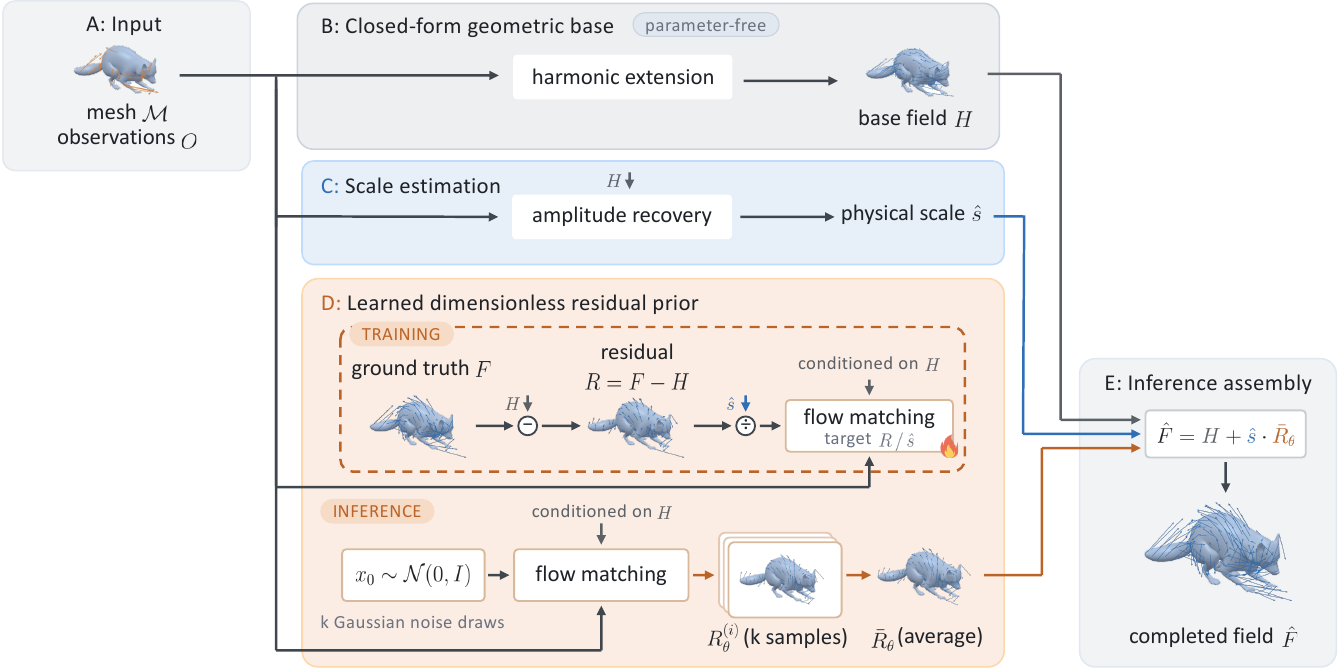}
\vspace{-6mm}
\caption{\textbf{\textsc{Cepheid} constructs dense motion by decoupling residual structure from physical scale.} Given a mesh $\mathcal{M}$ with coordinates $X$ and sparse observed displacements $O$ on vertex subset $\Omega$ (A), harmonic extension first yields a closed-form base field $H$ (B). Amplitude recovery then estimates physical scale $\hat{s}$ using frozen fitted coefficients (C; Algorithm~\ref{alg:amplitude_recovery}). Conditioned on $H$ and $O$, the flow-matching network learns the normalized target $R/\hat{s}$ with $R = F - H$ during training, while mapping $k$ Gaussian noise draws to samples averaged into $\bar{R}_\theta$ at inference (D); it is the sole component trained by gradient descent (flame icon). Finally, the completed field is assembled as $\hat{F} = H + \hat{s} \cdot \bar{R}_\theta$ (E).}
\label{fig:method}
\end{figure}

Let $\mathcal{M} = (\mathcal{V}, \mathcal{E})$ be a surface mesh with $N = |\mathcal{V}|$ vertices and edge set $\mathcal{E}$, and let $X \in \mathbb{R}^{N \times 3}$ denote its source-frame coordinates (Figure~\ref{fig:method}). Vertex displacements yield a dense motion field $F \in \mathbb{R}^{N \times 3}$ whose $i$-th row is the 3D translation of vertex $v_i$ between two frames. Motion is observed solely on a subset $\Omega \subset \mathcal{V}$ ($M = |\Omega|$), giving sparse observations $O = F_\Omega \in \mathbb{R}^{M \times 3}$. We recover $F$ on the remaining $N - M$ unobserved vertices from $X$, $\Omega$, and $O$ alone. Density $M/N$ spans 0.5\% to 50\%, assuming no skeletons, registration, or cross-mesh correspondence.

The task admits a closed-form geometric prior, so learning only needs to model what that analytical prior leaves behind. Let $\mathbf{L} = \mathbf{D} - \mathbf{A} \in \mathbb{R}^{N \times N}$ be the unweighted graph Laplacian of $\mathcal{E}$, where $\mathbf{A}$ is the adjacency matrix and $\mathbf{D}$ is the degree matrix. The discrete harmonic extension of the observed displacements is the solution to the Dirichlet problem on the mesh graph:
\begin{equation} \small
    (\mathbf{L} H)_i = 0 \quad \forall v_i \notin \Omega, \qquad \text{subject to} \quad H_\Omega = O,
    \label{eq:dirichlet_solve}
\end{equation}
requiring one sparse linear solve per observation condition. We denote this solve as $\mathcal{H}$ and the resulting dense base field as $H = \mathcal{H}(\mathcal{E}, \Omega, O) \in \mathbb{R}^{N \times 3}$. This linear extension provides the simplest analytical prior in our setting, in the spirit of classical handle-based deformation such as Laplacian mesh editing \citep{LaplacianMeshEditing:2004} and as-rigid-as-possible modeling \citep{ARAP_modeling:2007}.

Operating strictly on mesh connectivity without metric edge lengths, templates, or correspondence, $\mathcal{H}$ serves as a parameter-free control across all geometry families. The residual motion to predict is:
\begin{equation} \small
    R = F - H \in \mathbb{R}^{N \times 3}.
    \label{eq:residual_target}
\end{equation}

\vspace{-1mm}
\subsection{Separating Residual Structure from Physical Scale}
\label{sec:dimensionless_residual}
\vspace{-1mm}

The extent-normalized residual amplitude varies by $18\text{--}39\times$ across training sources at matched density and drops by more than an order of magnitude across densities (Appendix~\ref{app:amplitude_rule}), motivating factoring physical amplitude out of learning targets.
Let $\mathrm{std}_{99.5}(\cdot)$ denote the empirical standard deviation over all scalar components ($3M$ for $O$) clipped to $\pm$ the 99.5th percentile of absolute values, guarded by $+10^{-12}$. Because target motion is unknown at test time, residual dispersion $s = \mathrm{std}_{99.5}(R)$ cannot be measured. Instead, amplitude recovery estimates physical scale $\hat{s}$ without requiring target motion (\S\ref{sec:amplitude_recovery}), yielding normalized target:
\begin{equation} \small
    R_{\text{train}} = \frac{R}{\hat{s}},
    \label{eq:rtrain}
\end{equation}
with dispersion $s / \hat{s}$. At inference, physical scale is restored via:
\begin{equation} \small
    \hat{F} = H + \hat{s} \bar{R}_\theta, \qquad \text{where} \quad \bar{R}_\theta = \frac{1}{k} \sum_{i=1}^k R_\theta^{(i)},
    \label{eq:inference_f}
\end{equation}
with $k=8$ draws. Only $\hat{s}$ is used at training and deployment; $s$ and oracle $s^*$ (\S\ref{sec:amplitude_eval}) serve strictly as analysis references.

\vspace{-1mm}
\subsection{Target-Independent Amplitude Recovery}
\label{sec:amplitude_recovery}
\vspace{-1mm}

Amplitude recovery multiplies a dimensional gauge by a dimensionless correction:
\begin{equation} \small
    \hat{s} = \sigma_{\text{obs}} \exp\left(b + w^\top z(\varphi)\right),
    \label{eq:amplitude_rule}
\end{equation}
where the gauge $\sigma_{\text{obs}} = \mathrm{std}_{99.5}(O)$ provides physical units $[L]$ and $z(\cdot)$ standardizes $\varphi \in \mathbb{R}^{11}$, containing dimensionless descriptors of observation density, graph connectivity to $\Omega$, observed motion variation, and base field properties and ratios to observations (Appendix~\ref{app:features}, Table~\ref{tab:features}).

Parameters $\{b, w\}$ are fitted via ridge regression on $\log(s / \sigma_{\text{obs}})$ on DT4D, AMASS, and Franka (Table~\ref{tab:coefficients}, Appendix~\ref{app:coefficients}) and frozen before held-out evaluation; Algorithm~\ref{alg:amplitude_recovery} (Appendix~\ref{app:fitting}) details inference, including clamping $\varphi$ to its training support. All four transfer-matrix rows share this rule, isolating differences to learned residual priors. The gauge $\sigma_{\text{obs}}$ collapses cross-source scale spread from $37.5\times$ to $3.2\times$ at the sparsest rung across held-out conditions of the three training sources, while the correction brings completion error within 1--8\% of the target-informed oracle at every density on cloth (Table~\ref{tab:amplitude}).

\vspace{-1mm}
\subsection{Residual Predictor and Generation}
\label{sec:backbone}
\vspace{-1mm}

The residual distribution $R_\theta$ is modeled with an MGN-T backbone \citep{iparraguirre2026meshgraphnettransformerscalablemeshbasedlearned}, placing a Transformer attention block \citep{vaswani2017attention} between two message-passing blocks \citep{gilmer2017neural, pfaff2021learning, grigorev2023hood}; lacking official code, we implement MGN-T ourselves, pricing attention against message passing in Appendix~\ref{app:attention}. Conditioned on geometry $X$, mask $\Omega$, displacements $O$, and base field $H$, messages pass over a 12-nearest-neighbor coordinate graph (mesh connectivity reaches the network only through $H$). We train with OT-CFM \citep{tong2024improving} and sample 30 Euler steps, averaging $k=8$ draws (error is essentially flat by $k=8$; Table~\ref{tab:ksweep}, Appendix~\ref{app:k_sweep}).

\vspace{-1mm}
\section{Experimental Setup}

\vspace{-1mm}
\paragraph{Sources.} DT4D (articulated animals and humanoids), AMASS \citep{AMASS:ICCV:2019} (human motion capture), Franka (a rigid kinematic chain), and two held out of training entirely: cloth and a set of procedurally generated legged bodies (splits and citations in Appendix~\ref{app:data_sources}).

\vspace{-1mm}
\paragraph{Baselines and control.} We evaluate against three references: (1) Harmonic extension: a closed-form Dirichlet solve on the mesh graph (Eq.~\ref{eq:dirichlet_solve}) with zero learned parameters, serving as our parameter-free control rather than a competing method; (2) NSDP~\citep{tang2022neural}: regresses dense motion from sparse handles via a learned neural field, retrained on our standardized split; and (3) DOMA~\citep{DOMA}: optimizes a continuous per-scene deformation field from point trajectories without offline training.

\vspace{-1mm}
\paragraph{Protocol.} Table~\ref{tab:main} evaluates across five densities from 0.5\% to 50\% (7 held-out identities, 30 pairs each, 2 masks per pair, 2,100 conditions total). Table~\ref{tab:transfer} presents a transfer matrix where columns draw random masks (four per pair and density) on their own grids, reporting errors relative to column mean motion magnitude (Appendix~\ref{app:per_column_sampling}, Table~\ref{tab:protocol_per_column}). AMASS and DT4D withhold identities, Franka tests unseen trajectories, and cloth and legged bodies are held out at source level. Their cells are not interchangeable: Table~\ref{tab:main} balances pairs evenly across identities, whereas Table~\ref{tab:transfer} pools unevenly. Implementation details are in Appendix~\ref{app:implementation}. Frozen amplitude recovery (\S\ref{sec:amplitude_recovery}), fitted on DT4D, AMASS, and Franka, is applied by \emph{ours (unified)} in Table~\ref{tab:main}, all four Table~\ref{tab:transfer} priors, and all ablations keeping it.

\begin{table}[t]
\vspace{-4mm}
\centering
\small
\caption{\textbf{Baseline comparison on DT4D: NSDP and DOMA against ours, on one split.}
NSDP is above the no-learning base at every density; DOMA and ours are below it,
and ours is lowest at all five. Seven held-out identities, five densities (2,100 conditions), error divided by that density's mean motion magnitude. DOMA's row is its best iteration budget and reading at each density, chosen after scoring.}
\label{tab:main}
\begin{tabular}{lccccc}
\toprule
 & \multicolumn{5}{c}{Observed vertices} \\
\cmidrule(l){2-6}
err $\div$ mean $|F|$ & 0.5\% & 2\% & 5\% & 10\% & 50\% \\
\midrule
\emph{harmonic} & \emph{0.6255} & \emph{0.3533} & \emph{0.2257} & \emph{0.1598} & \emph{0.0754} \\
\midrule
NSDP \citep{tang2022neural}, retrained & 1.0406 & 0.5789 & 0.4018 & 0.3275 & 0.2683 \\
DOMA \citep{DOMA} & 0.5750 & 0.2370 & 0.1204 & 0.0720 & 0.0246 \\
\textbf{ours (unified)} & \textbf{0.3533} & \textbf{0.1116} & \textbf{0.0485} & \textbf{0.0300} & \textbf{0.0180} \\
\bottomrule
\end{tabular}
\end{table}

\begin{table}[t]
\centering
\small
\caption{\textbf{Transfer matrix: three single-family priors and one trained on all three,
each scored on five families.} Seven of the twelve off-diagonal cells sit at or
above their column's base, while the unified checkpoint is below it in every
column. 3,000 conditions per column, 3,840 on cloth; cells are completion error
divided by that column's mean motion magnitude, with \emph{harmonic} the
no-learning base. $\dagger$ The DT4D column's denominator is in mAU, that dataset's own artist units, not a length.
$\ddagger$ No arm trains on cloth or on the legged bodies.}
\label{tab:transfer}
\begin{tabular}{lccc;{2pt/2pt}cc}
\toprule
 & \multicolumn{5}{c}{Evaluated on} \\
\cmidrule(l){2-6}
Model & DT4D $\dagger$ & AMASS & Franka & cloth $\ddagger$ & legged $\ddagger$ \\
\midrule
\emph{harmonic} & \emph{0.2685} & \emph{0.0693} & \emph{0.1011} & \emph{0.4377} & \emph{0.2474} \\
\midrule
ours (DT4D) & \textbf{0.1097} & 0.0698 & 0.0782 & 0.4256 & 0.2238 \\
ours (AMASS) & 0.2701 & \textbf{0.0182} & 0.0459 & 0.5493 & 0.2482 \\
ours (Franka) & 0.2910 & 0.0797 & \textbf{0.0155} & 0.4717 & 0.2426 \\
ours (unified) & 0.1243 & 0.0214 & 0.0201 & \textbf{0.3692} & \textbf{0.2225} \\
\bottomrule
\end{tabular}
\end{table}

\vspace{-1mm}
\paragraph{Metric.} Let $\mathcal{V}_{\text{eval}}^{(c)} \subseteq \mathcal{V} \setminus \Omega^{(c)}$ denote the evaluated unobserved vertices in condition $c$ (excluding fixed handle vertices on cloth). For each condition $c$, we compute the mean per-vertex $L_2$ completion error $\bar{e}_c = \frac{1}{|\mathcal{V}_{\text{eval}}^{(c)}|} \sum_{v \in \mathcal{V}_{\text{eval}}^{(c)}} \|F_v^{(c)} - \hat{F}_v^{(c)}\|_2$ and mean ground-truth displacement $\bar{f}_c = \frac{1}{|\mathcal{V}_{\text{eval}}^{(c)}|} \sum_{v \in \mathcal{V}_{\text{eval}}^{(c)}} \|F_v^{(c)}\|_2$, both in the source's stored coordinates. The metric is the pooled relative error:
\begin{equation} \small
    E = \frac{\sum_c u_c \bar{e}_c}{\sum_c u_c \bar{f}_c},
    \label{eq:pooled_metric}
\end{equation}
where the sums run over the conditions of one table cell and $u_c$ is the per-condition factor that converts stored coordinates to physical length: the bounding-box extent times millimeters per native unit on the three extent-normalized sources (DT4D, AMASS and cloth), and millimeters per native unit alone on Franka and the legged bodies, which are stored unnormalized (on DT4D the result is mAU, not a length). Averaging over $\mathcal{V}_{\text{eval}}^{(c)}$ removes vertex-count dependence, so each condition is weighted in the pooled quotient by its mean motion magnitude in physical units ($u_c \bar{f}_c$), not by its vertex count.

Learning gain is $G = 1 - E / E_{\text{base}}$ ($G > 0$ positive, $G < 0$ negative transfer). Result tables report relative completion error $E$ directly, using harmonic extension as $E_{\text{base}}$, except Tables~\ref{tab:connectivity_free} (condition mean error in mAU, $\frac{1}{C}\sum_c u_c \bar{e}_c$) and~\ref{tab:finetune} (Franka links 1--7 error in mm). Relative metric $E$ mitigates DT4D's $118\times$ cross-identity artist-unit disparity. Appendix~\ref{app:pooling} details why the metric pools rather than averages per-condition ratios, and the normalization used in Table~\ref{tab:transfer}.

\vspace{-1mm}
\section{Results}

\vspace{-1mm}
\subsection{Comparison with NSDP and DOMA on DT4D}
\label{sec:baselines}
\vspace{-1mm}

Our model achieves lower error than NSDP and DOMA at all five densities on DT4D, landing 44--81\% below harmonic extension (Table~\ref{tab:main}). All three methods are evaluated on the same seven held-out identities of one split: NSDP is retrained on that split's training identities, DOMA fits per scene and needs no training split, and ours is the unified checkpoint of Table~\ref{tab:transfer}.

The harmonic row is the parameter-free control: a cell above it is one where harmonic extension alone would have been more accurate. DOMA is below the base at every density, by 8\% at 0.5\% rising to 67\% at 50\%, with error 1.4--2.5$\times$ ours. NSDP is above the base at every density, at 1.6--3.6$\times$ the base's error.

\vspace{-1mm}
\subsection{Transferring across geometry families}
\label{sec:transfer}
\vspace{-1mm}

To test cross-family residual transfer, we compare three single-family priors (DT4D animals, AMASS human bodies, Franka arm) and one unified prior trained on all three. All four share architecture, training budget, and frozen amplitude recovery fitted on the training sources, differing only in training data. Table~\ref{tab:transfer} evaluates these priors on the training families and two source-level held-out families: cloth (a continuum sheet without parts or kinematic chains) and nine procedural legged bodies with 8 to 30 degrees of freedom. Columns are normalized by mean motion magnitude, making error comparable within a column, not across columns.

Column denominators, for recovering a length: DT4D 85.511 mAU, AMASS 127.369 mm, Franka 3.287 mm, cloth
114.454 mm, legged 3.365 mm.

Seven of twelve off-diagonal specialist cells land at or above harmonic extension. On cloth, the AMASS and Franka priors are 25.5\% and 7.8\% above the base; on legged bodies, all specialists are within 10\%, with the AMASS prior 0.3\% above. The unified model beats the base in pooled error across all five families, crossing above harmonic extension only at 0.5\% on legged bodies (Table~\ref{tab:legged_density}) and outperforming it from 1\% up. Relative to the base, it reduces error by 15.6\% on cloth and 10.1\% on legged bodies, topping the DT4D prior (the specialist with lowest error in both columns) by 12.9 and 0.5 percentage points. On training sources, error remains below base, though 13--30\% above native specialists. Figure~\ref{fig:cloth_2pct} (Appendix~\ref{app:transfer_detail}) illustrates cloth at 2\% observation.

Per-density readings, the per-cell diagonals and each arm's training exposure are in
Appendix~\ref{app:transfer_detail}; replacing harmonic extension on mesh edges with a thin-plate
spline on a k-NN graph leaves completion viable at a cost concentrated at sparse observation
(Appendix~\ref{app:connectivity_free}).

\vspace{-1mm}
\subsection{Amplitude recovery on cloth}
\label{sec:amplitude_eval}
\vspace{-1mm}

On cloth, amplitude recovery lands within 1.4--8.1\% of the target-informed oracle error across densities (Table~\ref{tab:amplitude}). The network is frozen, so non-harmonic rows scale identical outputs (Figure~\ref{fig:amplitude_field}, Appendix~\ref{app:amplitude_rule}); the target-informed oracle and borrowed lookup table require ground truth and are not deployable. Amplitude recovery is below the borrowed lookup table by 0.4\% and 0.1\% at 0.5\% and 1\%, and 5.6--26.4\% from 2\% up. It outperforms harmonic extension across densities (1.4\% at 0.5\% to 58\% at 50\%), whereas raw observed scale without correction consistently underperforms it.

\begin{table}[t]
\vspace{-4mm}
\centering
\small
\caption{\textbf{Amplitude rules on cloth: one set of frozen network predictions, scaled four ways, against the no-learning base.} Cloth is held out of training; each cell is error $\div$ mean $|F|$. \emph{Target-informed oracle}: for each condition, the single scalar that minimizes error along the residual the network predicted, chosen after seeing the target and not computable at inference. \emph{harmonic}: the base field $H$ alone, with no learned residual. \emph{Raw observed scale}: $\hat{s} = \sigma_{\mathrm{obs}}$, the dispersion of the observed displacements, without the fitted correction. \emph{Borrowed lookup table}: a per-density amplitude table fitted on one training source, taking at each density whichever of the three gives the lowest error on cloth. \emph{Amplitude recovery}: $\sigma_{\mathrm{obs}}$ times the correction fitted on the training sources (Algorithm~\ref{alg:amplitude_recovery}), within 1--8\% of the oracle at every density.}
\label{tab:amplitude}
\begin{tabular}{lccccc}
\toprule
 & \multicolumn{5}{c}{Observed vertices} \\
\cmidrule(l){2-6}
cloth, err $\div$ mean $|F|$ & 0.5\% & 1\% & 2\% & 10\% & 50\% \\
\midrule
\emph{Target-informed oracle} & \emph{0.6283} & \emph{0.5175} & \emph{0.3917} & \emph{0.1492} & \emph{0.0529} \\
\midrule
\emph{harmonic} & \emph{0.6888} & \emph{0.5925} & \emph{0.4953} & \emph{0.2824} & \emph{0.1295} \\
\midrule
Raw observed scale (no correction) & 0.8073 & 0.7326 & 0.7650 & 1.1599 & 1.3197 \\
\emph{Borrowed lookup table (best source)} & \emph{0.6817} & \emph{0.5530} & \emph{0.4330} & \emph{0.2056} & \emph{0.0652} \\
\textbf{Amplitude recovery (ours)} & \textbf{0.6790} & \textbf{0.5524} & \textbf{0.4087} & \textbf{0.1513} & \textbf{0.0548} \\
\bottomrule
\end{tabular}
\end{table}

The \emph{Borrowed lookup table (best source)} row takes, at each density, whichever training table minimizes cloth error; because the optimal source varies across densities, hindsight selects no single table. The \emph{Target-informed oracle} row is the per-condition scalar $s^{*} = \arg\min_\alpha \sum_{v \in \mathcal{V}_{\mathrm{eval}}^{(c)}} \|(H + \alpha \bar{R}_\theta)_v - F_v\|_2$, the optimal scalar achievable for the network's predicted residual.

\vspace{-1mm}
\subsection{Ablations}
\vspace{-1mm}

We ablate \textsc{Cepheid}'s five design choices by removing each from the unified model (\emph{ours (unified)} in Table~\ref{tab:transfer}), retraining when necessary and scoring across the five columns. Removing harmonic extension eliminates its conditioning path; disabling amplitude recovery sets $\hat{s} = 1$. Table~\ref{tab:ablation} reports held-out families, and Table~\ref{tab:ablation_train} (Appendix~\ref{app:transfer_detail}) reports the three training families.

\begin{table}[t]
\centering
\small
\caption{\textbf{Component ablation on the two held-out families: each row removes one
choice from \textsc{Cepheid}.} Two removals can turn learning into negative
transfer; three cannot. A check marks a component the row keeps; in the amplitude column a dash is $\hat{s} = 1$, and in the samples
column a check is eight samples and a dash is one. The three training columns are
Table~\ref{tab:ablation_train}.}
\label{tab:ablation}
\setlength{\tabcolsep}{3pt}
\begin{tabular}{lccccc@{\hspace{10pt}}cc}
\toprule
 & \tl{harmonic}{\strut} & \tl{mean of}{$k$ samples} & \tl{flow}{matching} & \tl{conditioned}{on harmonic} & \tl{amplitude}{recovery} & cloth $\ddagger$ & legged $\ddagger$ \\
\midrule
\emph{harmonic} & & & & & & \emph{0.438} & \emph{0.247} \\
\midrule
\multirow{6}{*}{\begin{tabular}[c]{@{}l@{}}\textsc{Cepheid}\\(ours)\end{tabular}} & \no & \yes & \yes & \no & \yes & 1.047 & 0.585 \\
 & \yes & \no & \yes & \yes & \yes & 0.416 & 0.240 \\
 & \yes & \yes & \no & \yes & \yes & 0.415 & 0.226 \\
 & \yes & \yes & \yes & \no & \yes & 0.400 & 0.244 \\
 & \yes & \yes & \yes & \yes & \no & 0.391 & 0.409 \\
 & \yes & \yes & \yes & \yes & \yes & \textbf{0.369} & \textbf{0.222} \\
\bottomrule
\end{tabular}
\end{table}

\textbf{Two removals can cross the base.} Without harmonic extension, the model is above it in both
columns, by 2.4$\times$ on each; without amplitude recovery, it crosses on legged and stays below
on cloth. The remaining three removals stay below harmonic extension in both columns and therefore
never cross it. Reading the same weights at $k = 1$ is an inference setting rather than a trained arm,
and costs 8--13\% on the two held-out families.

The ablation clarifies component roles. Harmonic extension and amplitude recovery are the only removals that can turn learning into negative transfer relative to the no-learning base: harmonic extension supplies the geometric base field, while amplitude recovery establishes physical scale without target motion. Conditioning, flow matching, and multi-sampling affect gain magnitude without determining its sign in these experiments.

\vspace{-1mm}
\section{Applications}
\label{sec:applications}
\vspace{-1mm}

\begin{figure}[t]
\vspace{-3mm}
\centering
\includegraphics[width=\textwidth]{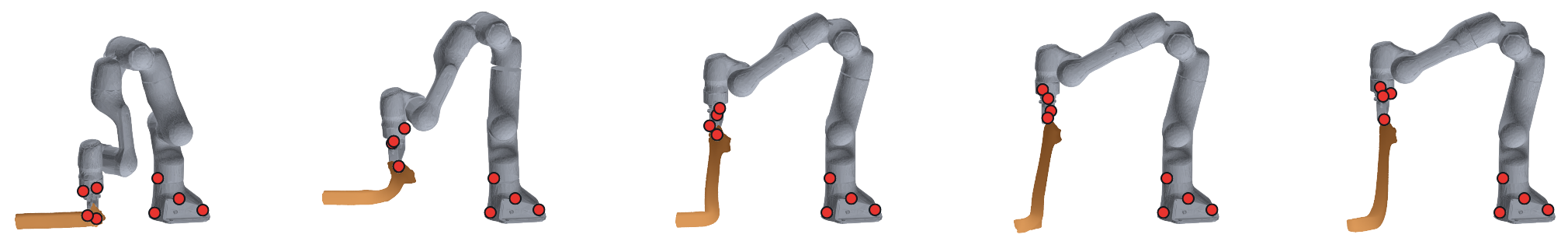}
\vspace{-7mm}
\caption{\textbf{Demo: five independent completions on a multi-body point set in Isaac Lab \citep{mittal2025isaaclab}.} The arm and grasped body form a face-free point set coupled by a 12-NN graph. From eight proprioceptive points (red; zero on object; markers can overlap at this viewpoint), the model predicts plausible coupled motion.}
\label{fig:robot_lift}
\end{figure}

\begin{figure}[t]
\centering
\includegraphics[width=\textwidth]{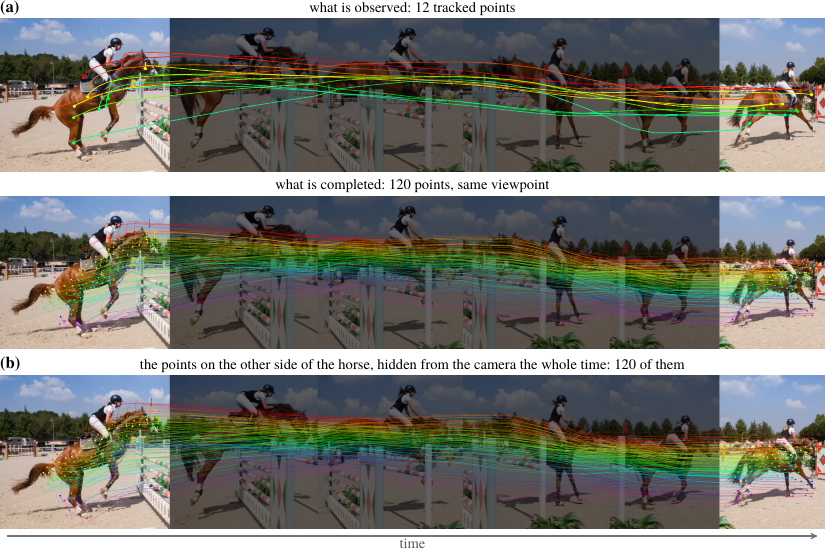}
\vspace{-7mm}
\caption{\textbf{Demo: densifying video point tracks.} Across frames 4--39, the model observes the same 12 tracked vertices in every frame. (a) Twelve observed points and 120 completed points. (b) 120 camera-opposed far-side points (0 of 2,164 ever observed). Without ground truth, this shows the checkpoint connects to a tracker and completions exist on the far side, not that either is accurate. Overlays degrade away from fitted frame 18 (Appendix~\ref{app:farside}).}
\label{fig:horsejump_strip}
\end{figure}

In two settings the observation comes from a signal that exists independently of the model: a robot's proprioception, and a point tracker's output. In Isaac Lab \citep{mittal2025isaaclab}, a Franka arm grips and lifts a soft bar (Figure~\ref{fig:robot_lift}). We represent the arm, base, and payload as a unified 5,525-point set without faces, connected solely by a 12-nearest-neighbor spatial graph without contact constraints. Observations comprise eight robot proprioceptive points (four on the gripper, four on the base) and zero on the payload. Figure~\ref{fig:robot_lift} shows five independent completions, propagating boundary signals across bodies without kinematic models. On a DAVIS \citep{Perazzi2016} sequence, 12 consistent vertex tracks derived from CoTracker3 \citep{karaev2024cotracker3} across frames 4--39 serve as sparse inputs on the target. These tracker outputs, rather than exact displacements, are completed across a mesh reconstructed from one frame by TripoSR \citep{TripoSR2024}. Figure~\ref{fig:horsejump_strip}b visualizes 120 completed vertices on the animal's far side, facing away from the reference camera and disjoint from the observed set (Appendix~\ref{app:farside}).

\vspace{-1mm}
\section{Conclusion}
\label{sec:conclusion}
\vspace{-1mm}

\textsc{Cepheid} completes dense motion across geometry families without skeletons, registration, or cross-mesh correspondence. Discrete harmonic extension is our parameter-free base and control; seven of twelve single-family transfers fail to beat it. Decoupling residual structure from scale, we estimate amplitude without target motion alongside a dimensionless flow-matching prior; on unseen cloth, completion error is within 1--8\% of a target-informed oracle. One unified model beats harmonic extension on all five families (15.6\% lower error on cloth, 10.1\% on legged bodies) and applies without retraining to point-tracked animal video and a robot grasping a soft body.

\vspace{-1mm}
\paragraph{Limitations.} Our quantitative evaluation assumes exact vertex displacements, free of sensor noise, lost tracks, depth distortion, or scanned geometry, and only AMASS and the RH20T episodes in the appendix are recorded data: DT4D is artist-animated, Franka uses forward kinematics on random joint trajectories, and cloth and legged robots are simulated. Training also uses random masks only; the prior transfers zero-shot to a fixed gripper placement (Appendix~\ref{app:finetune}), but fixed three-marker layouts gain substantially from fine-tuning (Table~\ref{tab:finetune}). Validation on noisy inputs and training across mask families are left to future work.

\label{endofmaintext}

\bibliography{iclr2027_conference}
\bibliographystyle{iclr2027_conference}

\appendix

\section{Implementation details}
\label{app:implementation}

\subsection{Network input}

Initial node features comprise 53 channels: the flow-matching state $x_t$ (3); the observed displacements $O$ (3), set to zero at unobserved vertices; a binary observation indicator (1); a Fourier position encoding of coordinates \citep{mildenhall2020nerf} centered and divided by the bounding-box extent, using frequencies $2^k \pi$ for $k \in \{0, \dots, 5\}$ with sine and cosine per coordinate (36); the mean and range (maximum minus minimum) of the observed displacements across observed vertices ($3 + 3$) and the observed fraction (1), replicated across all vertices; and the harmonic extension $H$ (3). The predicted amplitude $\hat{s}$ from amplitude recovery is not a network input; the network output is multiplied by $\hat{s}$ after sampling, and the training target is the residual $R$ divided by $\hat{s}$. The continuous flow time $t$ is embedded and added to node representations after the input encoder rather than concatenated.

\subsection{Training}

All models are trained using the Adam optimizer with PyTorch default settings ($\beta_1 = 0.9$, $\beta_2 = 0.999$, $\epsilon = 10^{-8}$, and no weight decay), with a learning rate of $3 \times 10^{-4}$ following a 1,000-step linear warmup and a cosine decay to 0.02 of that rate. Training uses gradient clipping at a maximum $\ell_2$ norm of 1.0, a batch size of 16, and single-precision floating point (fp32) on a single NVIDIA RTX PRO 6000 Blackwell GPU (96 GB VRAM) per run with random seed 0 for 500,000 steps. Peak memory is dominated by the largest training meshes (AMASS, 6,890 vertices), which consume approximately 5.3 GB of VRAM per sample. Every arm in Tables~\ref{tab:transfer}, \ref{tab:ablation}, and~\ref{tab:ablation_train} uses these values.

\subsection{Sampling}

We use OT-CFM \citep{tong2024improving}. At inference, probability trajectories are integrated using 30 uniform Euler steps, and predictions are averaged across $k = 8$ independent stochastic draws. Each completion requires one sparse linear solve for $H$, one evaluation of the amplitude rule, and 30 Euler integration steps, where the $k = 8$ sample draws are evaluated simultaneously by batching 8 disjoint graph copies, executing the network 30 times on $8N$ nodes per completion.

\subsection{Source sampling for the unified model}

Training batches for the unified model are drawn across sources according to the fixed ratio: DT4D 0.5, AMASS 0.3, and Franka 0.2.

\subsection{Data sources and splits}
\label{app:data_sources}

\begin{itemize}
\item \textbf{DT4D} \citep{li20214dcomplete}: articulated animals and humanoids. Training uses
116 identities; seven identities no arm trains on (elkML, foxYSY, ironman, moose45D, mremireh,
raccoonVGG, tigerD8H) are held out, and Tables~\ref{tab:main} and~\ref{tab:transfer} evaluate on them.
\item \textbf{AMASS} \citep{AMASS:ICCV:2019}: human motion capture under the official split.
Training uses the nine official-train sub-datasets, ACCAD \citep{AMASS_ACCAD}, BMLrub
\citep{AMASS_BMLrub}, CMU \citep{AMASS_CMU}, EKUT and KIT
\citep{AMASS_KIT-CNRS-EKUT-WEIZMANN,AMASS_KIT-CNRS-EKUT-WEIZMANN-2}, Eyes Japan
\citep{AMASS_EyesJapanDataset}, PosePrior \citep{AMASS_PosePrior}, TCDHands
\citep{AMASS_TCDHands} and TotalCapture \citep{AMASS_TotalCapture}. The four official-validation
sub-datasets, HDM05 \citep{AMASS_HDM05}, HumanEva \citep{AMASS_HumanEva}, MoSh
\citep{AMASS_MoSh} and SFU \citep{AMASS_SFU}, are held out and form the AMASS column of
Table~\ref{tab:transfer}.
\item \textbf{Franka}: forward-kinematic trajectories of the Panda arm, generated from its URDF
with random joint-space splines; 54 trajectories train, and 12 held out form its column.
\item \textbf{Cloth}: the flag data of MeshGraphNets \citep{pfaff2021learning}. No arm trains on
it; Table~\ref{tab:transfer} evaluates 24 trajectories of its validation split.
\item \textbf{Legged}: nine GenBot1K embodiments \citep{ai2025towards}, three quadrupeds, three
hexapods and three humanoids, moving under the released locomotion experts rolled out in Isaac Lab
\citep{mittal2025isaaclab}. No arm trains on them.
\item \textbf{Commercial manipulators}: one RH20T episode each \citep{fang2024rh20t} for a Flexiv,
a Kuka and a UR5 (Appendix~\ref{app:transfer_detail}); no arm trains on them.
\end{itemize}

\subsection{Shared experimental protocol across arms}

The four arms share their architecture, base conditioning, flow-matching objective, optimizer, schedule, total training budget (500,000 steps), and random-mask family. The arms differ in the sources their residual prior was trained on and in nothing else.

$\mathbf{L}$ uses unit edge weights, so it is defined on any graph, including the face-free 12-nearest-neighbor graph of Section~6, on which cotangent weights are undefined \citep{pinkall1993computing}. The harmonic solve adds $10^{-6}\mathbf{I}$ (Tikhonov regularization) to $\mathbf{L}_{uu}$, so a connected component with no observed vertex receives $H = 0$; this occurs on DT4D at 0.5\% observation. In Section~6, a 12-nearest-neighbor graph is constructed once from the initial frame and reused across all five frames for the unweighted Laplacian solve yielding $H$ and for computing graph-hop features, whereas the network's 12-nearest-neighbor connectivity is reconstructed at each frame.

The committed configurations record the learning-rate schedule, clipping, batch, step count and seed; Adam's betas, eps and weight decay are PyTorch's defaults, and the checkpoints store none of these values.

\subsection{Evaluation protocol per column}
\label{app:per_column_sampling}

\needspace{12\baselineskip}
\begin{table}[t]
\centering
\small
\caption{\textbf{What each evaluation column samples.} Each source is sampled on its own density grid. Its data are organized either by frame pair or by frame across continuous trajectories. Masks are drawn from the random-mask
family throughout, four per pair and density.}
\label{tab:protocol_per_column}
\begin{tabular}{llccl}
\toprule
column & sampled over & masks & densities & motion norm. \\
\midrule
DT4D & 250 pairs & 4 & 0.5 / 10 / 50\% & extent \\
AMASS & 250 pairs & 4 & 0.5 / 10 / 50\% & extent \\
Franka & 250 pairs & 4 & 0.5 / 10 / 50\% & none \\
cloth & 24 trajectories $\times$ 8 frames & 4 & 0.5 / 1 / 2 / 10 / 50\% & extent \\
legged & 250 pairs & 4 & 0.5 / 10 / 50\% & none \\
\bottomrule
\end{tabular}
\end{table}

Each cell of
Table~\ref{tab:transfer} pools over its own column's grid.

\section{Cross-family transfer and training-set ablations}
\label{app:transfer_detail}

\subsection{Reading the transfer matrix}

In raw physical units, the unified model achieves completion errors of $0.066\,\text{mm}$ on Franka and $10.6\,\text{mAU}$ on DT4D. Furthermore, the dataset-wide mean motion denominators span a $39\times$ factor across columns. This motivates normalizing each column by its mean ground-truth displacement in Table~\ref{tab:transfer}.

Dividing by one constant per column makes relative error directly readable within each column, but a normalized-only table would not be invertible back to a length. The column denominators are therefore printed: DT4D $85.511\,\text{mAU}$, AMASS $127.369\,\text{mm}$, Franka $3.287\,\text{mm}$, cloth $114.454\,\text{mm}$, and legged $3.365\,\text{mm}$, so every normalized cell in Table~\ref{tab:transfer} can be inverted back to physical length.

\subsection{The diagonals, and what each column holds out}

Every column in Table~\ref{tab:transfer} holds out evaluation data, but the nature of the held-out split differs across sources:
\begin{itemize}
    \item \textbf{AMASS} evaluates generalization across subjects by withholding four complete sub-datasets from training (Appendix~\ref{app:data_sources}).
    \item \textbf{DT4D} evaluates generalization across body shapes by withholding seven distinct geometric identities.
    \item \textbf{Franka} evaluates generalization across motions by withholding twelve continuous trajectories executed on a single robot geometry.
\end{itemize}
Consequently, the Franka diagonal benchmarks unseen trajectories on a seen physical embodiment, whereas the AMASS and DT4D diagonals benchmark unseen embodiments. The three diagonal evaluations are therefore structurally distinct.

\subsection{Training exposure}
\label{app:exposure}

At matched optimization steps ($500{,}000$) and batch size ($16$), each specialist model processes $8.0$ million training samples drawn exclusively from its single native domain. In comparison, the unified model processes $4.0$ million samples from DT4D, $2.4$ million from AMASS, and $1.6$ million from Franka ($2\text{--}5\times$ fewer exposures per individual source).

Moreover, the unified model sees no additional unique training items: it encounters the identical set of unique items on DT4D and Franka, and only 8\% of the unique training items available in AMASS. The performance of the unified model is therefore not explained by greater exposure to any one source.

\subsection{Emission ratios}
\label{app:emission}

What the unified model exhibits that the specialists do not is the ability to output a residual correction of approximately the required scale across diverse target geometries. We measure this property using the emission ratio, defined as the mean per-vertex $\ell_2$ length of the predicted physical correction $\hat{s} \bar{R}_\theta$ divided by that of the unobserved residual left by the base field $F - H$, over the evaluated vertices and pooled across conditions.

For the unified model, this ratio remains stably bounded between $0.57$ and $0.99$ across the four evaluation columns that record these statistics (the legged evaluation outputs do not store these field diagnostics), remaining below unity throughout. In contrast, specialist priors deviate substantially from unity off-source: the Franka prior yields emission ratios of $0.21\text{--}0.96$, under-predicting the required correction across all columns, while the AMASS prior yields ratios of $0.94\text{--}1.51$, over-predicting on most. Consequently, one specialist consistently under-corrects while the other over-corrects on the identical completion task.

This diagnostic also clarifies the unscaled ablation in Tables~\ref{tab:ablation} and~\ref{tab:ablation_train}, where amplitude recovery is removed ($\hat{s} = 1$). Across the density ladder, the emission ratio of the fixed-amplitude arm expands from $0.72$ to $1.03$ on DT4D, $0.77$ to $1.61$ on AMASS, and $1.09$ to $5.93$ on Franka, whereas the adaptive predictor maintains ratios near unity. The resulting performance penalties are $1.17\text{--}1.79\times$ on DT4D, $1.58\text{--}5.98\times$ on AMASS, $4.66\text{--}94.0\times$ on Franka, and $1.02\text{--}1.48\times$ on cloth. Franka represents the extreme case of this mechanism: when evaluated in the bounding-box-normalized coordinates where the amplitude acts, its residual is roughly $26\text{--}42\times$ smaller than DT4D's across the density grid, causing a single shared static amplitude to severely overshoot on that domain first.

\subsection{The three commercial manipulators, per arm}

We evaluate the unified checkpoint zero-shot on one recorded teleoperation episode each from three commercial manipulators in RH20T \citep{fang2024rh20t}: a Flexiv 7-DoF arm, a Kuka LBR iiwa, and a Universal Robots UR5. Geometries and kinematic topologies are parsed directly from vendor-shipped URDFs, with completion executed without any task-specific fine-tuning.

\begin{figure}[t]
\centering
\includegraphics[width=\textwidth]{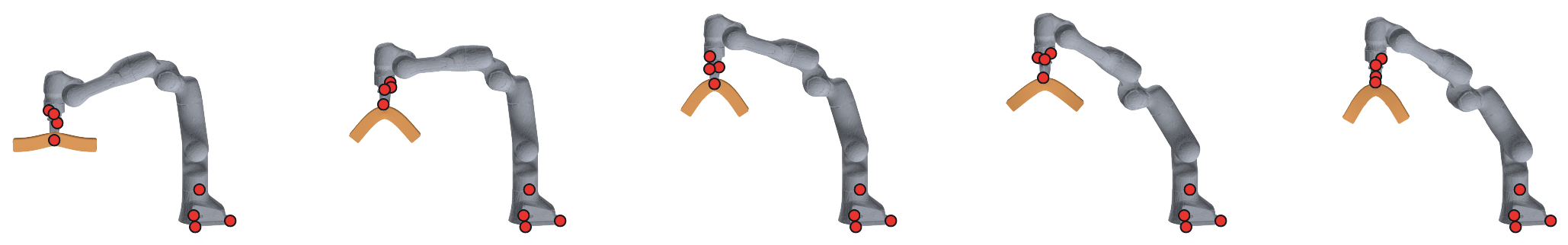}
\caption{\textbf{Demo: soft-bar manipulation in Newton.} Five instants of a Franka arm carrying a soft bar simulated in Newton 1.5.2 \citep{newton2025} as a face-free 12-NN point set. Given eight proprioceptive points (red) and none on the sagging bar, the frozen checkpoint completes dense motion, closely matching harmonic extension on the unobserved payload.}
\label{fig:newton_bar}
\end{figure}

Observations are restricted to four points on the end-effector and four points on the fixed base, corresponding to $0.200\%$ of the manipulator's vertices. Across all three systems, completions remain below harmonic extension, and in all cases the unified model outperforms the specialist prior trained exclusively on the Franka manipulator. Because these experiments evaluate single demonstration episodes ($24\text{--}25$ frame pairs each), we report directional validation across diverse robotic kinematic chains rather than aggregate benchmark statistics. Figure~\ref{fig:newton_bar} demonstrates a Franka arm carrying a gravity-sagged soft bar simulated in Newton 1.5.2 \citep{newton2025}; from eight robot points without retraining, the completion on the unobserved bar closely matches harmonic extension.

\needspace{18\baselineskip}
\subsection{Transfer to the held-out families by density}

\begin{table}[t]
\centering
\small
\caption{\textbf{Cloth by density: the four priors and the base at each rung.} Every
specialist crosses the no-learning base and the unified model never does. 3,840
conditions, 768 per density; Table~\ref{tab:transfer} pools all five rungs.}
\label{tab:cloth_density}
\begin{tabular}{lccccc}
\toprule
err $\div$ mean $|F|$, cloth & 0.5\% & 1\% & 2\% & 10\% & 50\% \\
\midrule
\emph{harmonic} & \emph{0.6888} & \emph{0.5925} & \emph{0.4953} & \emph{0.2824} & \emph{0.1295} \\
\midrule
DT4D & 0.8131 & 0.6479 & 0.4578 & 0.1550 & \textbf{0.0544} \\
AMASS & 0.9511 & 0.8393 & 0.6440 & 0.2360 & 0.0765 \\
Franka & 0.7594 & 0.6451 & 0.5268 & 0.2926 & 0.1346 \\
\textbf{\textsc{Cepheid}} & \textbf{0.6790} & \textbf{0.5524} & \textbf{0.4087} & \textbf{0.1513} & 0.0548 \\
\bottomrule
\end{tabular}
\end{table}

\begin{table}[t]
\centering
\small
\caption{\textbf{Legged by density: the four priors and the base at each rung.} At the
sparsest rung every arm is above the no-learning base; at 1\% only the AMASS
prior still is, and from 2\% up none is. Table~\ref{tab:transfer} pools three of these five rungs: 0.5, 10 and 50\%.}
\label{tab:legged_density}
\begin{tabular}{lccccc}
\toprule
err $\div$ mean $|F|$, legged & 0.5\% & 1\% & 2\% & 10\% & 50\% \\
\midrule
\emph{harmonic} & \emph{0.6033} & \emph{0.4228} & \emph{0.2640} & \emph{0.0954} & \emph{0.0435} \\
\midrule
DT4D & 0.6227 & 0.4181 & 0.2274 & 0.0383 & 0.0102 \\
AMASS & 0.6955 & 0.4293 & 0.2127 & \textbf{0.0308} & 0.0183 \\
Franka & 0.6626 & 0.4195 & 0.2268 & 0.0478 & 0.0175 \\
\textbf{\textsc{Cepheid}} & \textbf{0.6222} & \textbf{0.3990} & \textbf{0.2067} & 0.0355 & \textbf{0.0096} \\
\bottomrule
\end{tabular}
\end{table}

Each specialist prior exceeds harmonic extension in its column under sparse observations. On cloth (Table~\ref{tab:cloth_density}), the Franka prior remains above the base across all densities, the AMASS prior through $2\%$, and the DT4D prior through $1\%$. On legged bodies (Table~\ref{tab:legged_density}), all three specialists sit above the base at $0.5\%$, only the AMASS prior remains above at $1\%$, and from $2\%$ upward all three fall below.

The unified model improves over the base across all five evaluated densities on cloth. On legged bodies, it sits above the base at $0.5\%$ and below it from $1\%$ upward. Across both held-out domains, the unified model achieves the lowest completion error among the four learned models at the three sparsest rungs ($0.5\%$, $1\%$, and $2\%$).

The cloth entries in Table~\ref{tab:transfer} pool across all five rungs in Table~\ref{tab:cloth_density}, whereas the legged entries in Table~\ref{tab:transfer} pool across the $0.5\%$, $10\%$, and $50\%$ rungs (the density grid shared across the source columns). Each rung is normalized by its own mean motion magnitude: across the $0.5\%$, $1\%$, $2\%$, $10\%$, and $50\%$ densities, the cloth denominators are $114.4419$, $114.4454$, $114.4417$, $114.4664$, and $114.4751\,\text{mm}$, and the legged denominators are $3.3657$, $3.3659$, $3.3664$, $3.3656$, and $3.3648\,\text{mm}$.

\needspace{24\baselineskip}
\subsection{Qualitative comparison with NSDP and DOMA on cloth}

\begin{figure}[t]
\centering
\includegraphics[width=\textwidth]{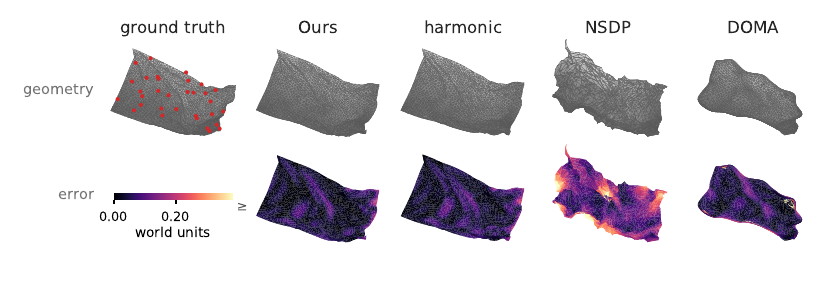}
\caption{\textbf{Qualitative comparison on cloth, a family no model trained on.}
Completion from 32 observed vertices of the 1,577 free vertices, shown as geometry above and
per-vertex error below on a shared ramp; the observed points are marked on the
ground truth. \emph{Ours} is the unified checkpoint and NSDP is retrained on our
DT4D split. NSDP tears the sheet and DOMA loses its shape and self-intersects,
while our completion keeps the flag intact and its error field dark. \textbf{Ours
is the only one of the three below harmonic extension here.}}
\label{fig:cloth_2pct}
\end{figure}

On an unseen cloth condition at 2\% observation, Figure~\ref{fig:cloth_2pct} compares our unified checkpoint, retrained NSDP, and per-scene DOMA against harmonic extension. Geometry is shown above and per-vertex error fields below.

\subsection{The ablation on the three training families}

\begin{table}[t]
\centering
\small
\caption{\textbf{The same component ablation on the three training families.} The arms,
conditions and scoring are those of Table~\ref{tab:ablation}, which carries the
two held-out columns; together they are one five-column ablation, with rows in that table's order (cloth, worst first). A check marks a component the row
keeps; in the amplitude column a dash is $\hat{s} = 1$, and in the samples column a check is eight samples and a dash is one.}
\label{tab:ablation_train}
\setlength{\tabcolsep}{3pt}
\begin{tabular}{lccccc@{\hspace{10pt}}ccc}
\toprule
 & \tl{harmonic}{\strut} & \tl{mean of}{$k$ samples} & \tl{flow}{matching} & \tl{conditioned}{on harmonic} & \tl{amplitude}{recovery} & DT4D & AMASS & Franka \\
\midrule
\emph{harmonic} & & & & & & \emph{0.269} & \emph{0.069} & \emph{0.101} \\
\midrule
\multirow{6}{*}{\begin{tabular}[c]{@{}l@{}}\textsc{Cepheid}\\(ours)\end{tabular}} & \no & \yes & \yes & \no & \yes & 0.752 & 0.341 & 0.457 \\
 & \yes & \no & \yes & \yes & \yes & 0.152 & 0.025 & 0.024 \\
 & \yes & \yes & \no & \yes & \yes & 0.126 & \textbf{0.021} & 0.028 \\
 & \yes & \yes & \yes & \no & \yes & 0.166 & 0.027 & 0.032 \\
 & \yes & \yes & \yes & \yes & \no & 0.152 & 0.038 & 0.209 \\
 & \yes & \yes & \yes & \yes & \yes & \textbf{0.124} & \textbf{0.021} & \textbf{0.020} \\
\bottomrule
\end{tabular}
\end{table}

Table~\ref{tab:ablation_train} presents the corresponding five-choice ablation evaluated on the three native training families. In the arm without harmonic extension, $H = 0$ leaves the four amplitude features computed from $H$ undefined; a $10^{-12}$ guard keeps each finite but far below its fitted range, so the support clamp fixes each at the lower edge. Removing harmonic extension causes error to exceed the harmonic row across all columns in both tables, by $4.9\times$ on AMASS. Removing amplitude recovery crosses the base on Franka ($0.020 \rightarrow 0.209$) while remaining below it on AMASS and DT4D; together with the legged column in Table~\ref{tab:ablation}, these represent the two columns out of five where removing amplitude recovery induces negative transfer.

The relative degradation from removing intermediate components does not strictly track the distinction between training and held-out domains: removing base conditioning degrades completion by 8.4\% relative to \textsc{Cepheid} on cloth, while removing flow matching costs 12.5\% and removing sample averaging costs 12.7\%. On AMASS, the corresponding removals degrade performance by 28.6\%, leave error unchanged at three decimals, and degrade it by 19.0\%, respectively.

Finally, evaluating single-draw inference ($k=1$) increases error by 19--23\% across the three training sources, compared to the 8--13\% increase observed on the two held-out families in Table~\ref{tab:ablation}.

\section{Amplitude recovery, backbone attention, and inference sampling}
\label{app:amplitude_rule}

\subsection{What is fitted}

We define the residual amplitude $s$ and the observed-displacement gauge $\sigma_{\mathrm{obs}}$ (\S\ref{sec:dimensionless_residual}) as:
\begin{equation}
    s = \mathrm{std}_{99.5}(R), \qquad \sigma_{\mathrm{obs}} = \mathrm{std}_{99.5}(O).
\end{equation}
Across the three training sources at matched density, $s$ spans $18\text{--}39\times$ when pooled per density over training data with extent-normalized objects (Franka is divided by its bounding-box extent only for this comparison). It also falls by another order of magnitude across the density grid while the ground-truth motion stays flat. Measured across DT4D training runs differing only in the base, the 99.5\%-clipped standard deviation of motion with the base set to zero varies by 6.8\% across 0.5\% to 50\% observation (max over min), whereas the same statistic of the residual after harmonic extension varies by $18.3\times$.

Because $\sigma_{\mathrm{obs}}$ is evaluated directly on observed displacements at inference, the ratio $s / \sigma_{\mathrm{obs}}$ is dimensionless. We fit a ridge regression model \citep{hoerl1970ridge} on its logarithm:
\begin{equation}
    \log\frac{s}{\sigma_{\mathrm{obs}}} \approx b + w^\top z(\varphi),
\end{equation}
where $z(\cdot)$ standardizes the feature vector $\varphi$ against frozen statistics. Exponentiating yields the served scale:
\begin{equation}
    \hat{s} = \sigma_{\mathrm{obs}} \exp\bigl(b + w^\top z(\varphi)\bigr).
\end{equation}
Fitting $\log(s / \sigma_{\mathrm{obs}})$ rather than $\log s$ directly removes dataset-specific unit conventions from the regression objective. Training targets $R / \hat{s}$ are clipped to $[-8, 8]$; inference evaluates $\hat{F} = H + \hat{s}\bar{R}_\theta$ by multiplication alone, so stationary observations yield $\hat{s} \approx 0$ and $\hat{F} = H$.

\textbf{The gauge does most of the unit work and the correction does the rest.}
Evaluated on held-out conditions of the three training sources with Franka in meters, the cross-source span of the geometric mean of per-condition $s$ at the sparsest rung is 37.5$\times$; the same geometric mean for per-condition $s / \sigma_{\mathrm{obs}}$ spans 3.2$\times$. What
the correction adds is variation that depends on the condition: the gauge alone is above the
no-learning base at every density on cloth, 0.8073 to 1.3197 against the base's 0.6888 to 0.1295
(Table~\ref{tab:amplitude}).

\begin{figure}[t]
\centering
\includegraphics[width=\textwidth]{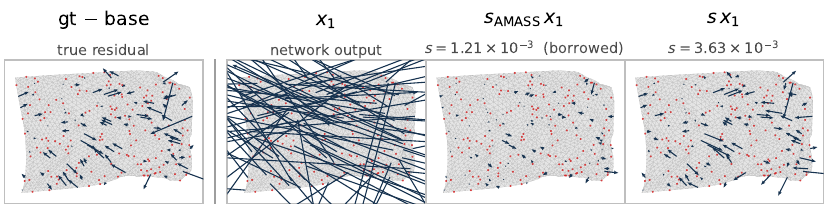}
\caption{\textbf{The predicted amplitude recovers the residual; a borrowed one does not.}
Cloth at 10\% observation, one shared arrow scale; red dots are the observed
vertices. The first panel is the true residual; the three that follow show the
same frozen network output, differing only in the scale applied to it. The
network's raw output carries no physical scale, and AMASS's table restores it at
a third of the predicted scale.}
\label{fig:amplitude_field}
\end{figure}

\subsection{The features}
\label{app:features}

The vector $\varphi \in \mathbb{R}^{11}$ comprises eleven entries, defined in design-matrix order in Table~\ref{tab:features}. We denote the observed displacements as $O$, the harmonic extension as $H$, vertex count as $N$, observed count as $M$, and mean vertex degree as $\bar{d} = 2|\mathcal{E}|/N$.

\needspace{21\baselineskip}
\begin{table}[t]
\centering
\small
\caption{\textbf{The eleven amplitude-recovery features, as the code computes them.} Together they form $\varphi \in \mathbb{R}^{11}$.}
\label{tab:features}
\begin{tabular}{@{}lp{255pt}@{}}
\toprule
\textbf{Feature} & \textbf{Mathematical Definition} \\
\midrule
\texttt{log\_frac} & $\log(M / N)$ \\
\texttt{log\_n\_obs} & $\log M$ \\
\texttt{log\_n\_vert} & $\log N$ \\
\texttt{log\_hop\_mean} & $\log(\bar{h} + 1)$, with $\bar{h}$ the mean graph hops to $\Omega$ over unobserved vertices that can reach it \\
\texttt{log\_hop\_max} & $\log(h_{\max} + 1)$, with $h_{\max}$ the maximum graph hops to $\Omega$ over the same vertices \\
\texttt{unreach\_frac} & Share of the unobserved vertices with no path to $\Omega$ \\
\texttt{obs\_norm\_cv} & $\operatorname{std}_j \|O_j\|_2 \,/\, \operatorname{mean}_j \|O_j\|_2$ \\
\texttt{log\_base\_over\_obs} & $\log \left( \operatorname{mean}_{i \notin \Omega} \|H_i\|_2 \,/\, \operatorname{mean}_j \|O_j\|_2 \right)$ \\
\texttt{log\_sigbase\_over\_sigobs} & $\log \left( \mathrm{std}_{99.5}(H_{\mathcal{V} \setminus \Omega}) \,/\, \sigma_{\text{obs}} \right)$ \\
\texttt{log\_base\_rough} & $\log \left( \operatorname{mean}_{(i,j) \in \mathcal{E}} \|H_i - H_j\|_2 \,/\, \operatorname{mean}_{i \in \mathcal{V}} \|H_i\|_2 \right)$ \\
\texttt{log\_base\_rough\_hop} & $\log \left( \bar{d} \operatorname{mean}_{(i,j) \in \mathcal{E}} \|H_i - H_j\|_2 \,/\, \operatorname{mean}_{i \in \mathcal{V}} \|H_i\|_2 \right)$ \\
\bottomrule
\end{tabular}
\end{table}

Every feature in Table~\ref{tab:features} is dimensionless: each represents a count, a proportion, or a ratio between two matching lengths. Native length scales cancel out before regression, allowing a single rule fitted on three training sources to be deployed directly onto unseen geometries (\S\ref{sec:amplitude_recovery}).

\subsection{Fitting}
\label{app:fitting}

From an initial pool of 15 candidate features across five groups (density, connectivity, observed-motion variation, base field, and mesh), an inner leave-one-source-out selection drops at most one entire group, which for the deployed fold eliminated the four mesh features to yield the final 11. The ridge regression is fitted on log amplitude ratios with sample weights balanced across the three training sources. The intercept $b$ is unpenalized. Feature standardization and the 1st/99th-percentile clamp bounds are frozen statistics, not free parameters: 12 fitted values and 56 numbers frozen in total. Clamp bounds are applied to $\varphi$ prior to standardization, ensuring that features exceeding the training distribution support are evaluated at the boundary rather than extrapolated.

\begin{algorithm}[h]
\caption{Inference-Time Amplitude Recovery}
\label{alg:amplitude_recovery}
\small
\begin{algorithmic}[1]
\Require Frozen rule: standardized weights $w$ and intercept $b$ (Table~\ref{tab:coefficients}); feature means $\mu_\varphi$, standard deviations $\sigma_\varphi$ (with $\oslash$ denoting element-wise division), and clamp bounds $\varphi_{\min}, \varphi_{\max}$, frozen statistics released with the code (Appendix~\ref{app:fitting})
\Ensure Predicted physical scale $\hat{s}$
\State $\sigma_{\text{obs}} \gets \mathrm{std}_{99.5}(O)$
\State $\varphi \gets \mathrm{ExtractFeatures}(\mathcal{E}, \Omega, O, H)$ \Comment{Table~\ref{tab:features}}
\State $\varphi \gets \mathrm{clip}(\varphi, \varphi_{\min}, \varphi_{\max})$
\State $z \gets (\varphi - \mu_\varphi) \oslash \sigma_\varphi$
\State $\hat{s} \gets \sigma_{\text{obs}} \cdot \exp(b + w^\top z)$
\State \Return $\hat{s}$
\end{algorithmic}
\end{algorithm}

\subsection{Amplitude-recovery coefficients}
\label{app:coefficients}

\begin{table}[t]
\centering
\small
\caption{\textbf{The deployed standardized amplitude-recovery weights $w$ and intercept $b = -2.439$.} The rule is the leave-cloth-out fold, fitted on standardized features from DT4D, AMASS and Franka with the mesh group dropped; individual weights are not interpretable because \texttt{log\_frac} $=$ \texttt{log\_n\_obs} $-$ \texttt{log\_n\_vert} and \texttt{log\_base\_rough\_hop} $=$ \texttt{log\_base\_rough} $+ \log \bar{d}$ exactly (Table~\ref{tab:features}).}
\label{tab:coefficients}
\begin{tabular}{@{}lr@{}}
\toprule
feature & $w$ \\
\midrule
$\texttt{log\_frac}$ & $-0.202$ \\
$\texttt{log\_n\_obs}$ & $-0.237$ \\
$\texttt{log\_n\_vert}$ & $-0.173$ \\
$\texttt{log\_hop\_mean}$ & $+0.171$ \\
$\texttt{log\_hop\_max}$ & $+0.189$ \\
$\texttt{unreach\_frac}$ & $+0.075$ \\
$\texttt{obs\_norm\_cv}$ & $+0.011$ \\
$\texttt{log\_base\_over\_obs}$ & $-0.057$ \\
$\texttt{log\_sigbase\_over\_sigobs}$ & $-0.068$ \\
$\texttt{log\_base\_rough}$ & $+0.118$ \\
$\texttt{log\_base\_rough\_hop}$ & $+0.110$ \\
\midrule
intercept $b$ & $-2.439$ \\
\bottomrule
\end{tabular}
\end{table}

Table~\ref{tab:coefficients} presents the learned standardized weights $w$ and intercept $b$. At the training mean, all standardized features evaluate to zero ($z = 0$), reducing the correction factor $\hat{s}/\sigma_{\mathrm{obs}}$ to $\exp(b) = 0.0873$. Under average training conditions, the residual amplitude is therefore approximately one-eleventh of the dispersion of the observed displacements ($\hat{s} \approx 0.0873 \sigma_{\mathrm{obs}}$).

During production deployment, features are standardized and clamped using the released configuration parameters, which reproduce the exact served amplitude values.

\subsection{Selection}

Within each leave-one-source-out fold, an inner cross-validation loop across the three remaining training sources determines whether to retain all five feature groups or eliminate one. The held-out evaluation source is strictly excluded from both feature selection and coefficient estimation.

\subsection{The per-density optimal table}

As an additional target-informed reference, we compute an oracle table whose per-density entries minimize completion error on cloth, representing the performance ceiling for any single-value-per-density rule accessing the ground truth. Amplitude recovery approaches this oracle within approximately one percentage point at the two sparsest densities, and outperforms it by 2.8\% to 10.8\% from 2\% observation density upward (Table~\ref{tab:oracle_table}). This advantage arises because an empirical table assigns a static scalar per density, whereas amplitude recovery adapts dynamically to individual test conditions.

\begin{table}[t]
\centering
\small
\caption{\textbf{Oracle table comparison on held-out cloth.} Conditions match Table~\ref{tab:amplitude}, normalized by mean motion magnitude. Relative to the ground-truth-fitted oracle table ($\alpha^*$), amplitude recovery differs by $+1.04\%$ at 0.5\%, $-0.09\%$ at 1\%, and reduces error by 2.81\% to 10.77\% from 2\% upward.}
\label{tab:oracle_table}
\begin{tabular}{lccccc}
\toprule
cloth, err $\div$ mean $|F|$ & 0.5\% & 1\% & 2\% & 10\% & 50\% \\
\midrule
\emph{harmonic} & \emph{0.6888} & \emph{0.5925} & \emph{0.4953} & \emph{0.2824} & \emph{0.1295} \\
Amplitude recovery (ours) & \textbf{0.6790} & \textbf{0.5524} & \textbf{0.4087} & \textbf{0.1513} & \textbf{0.0548} \\
\emph{Oracle table} ($\alpha^{*}$) & \emph{0.6720} & \emph{0.5529} & \emph{0.4205} & \emph{0.1670} & \emph{0.0614} \\
\midrule
Relative difference vs $\alpha^{*}$ & $+1.04\%$ & $-0.09\%$ & $-2.81\%$ & $-9.40\%$ & $-10.77\%$ \\
\bottomrule
\end{tabular}
\end{table}

\subsection{The attention ablation}
\label{app:attention}

Message passing resolves local deformation shared across neighboring vertices, whereas non-local attention relates distant surface regions lacking short graph paths. As reported in Table~\ref{tab:attention}, comparing MGN-T against a matched-budget message-passing control yields relative error reductions of 7.5\% at 0.5\% observation and between 2.0\% and 3.3\% from 2\% through 50\% density. Pooled over the five densities, MGN-T achieves a 4.9\% error reduction and is lower on all seven held-out identities.

\begin{table}[t]
\centering
\small
\caption{\textbf{Attention ablation on DT4D.} Comparison of MGN-T against a matched-budget message-passing control across seven held-out identities at five densities (0.5, 2, 5, 10, 50\%). MGN-T achieves a 4.9\% pooled relative reduction and yields lower error on all seven identities.}
\label{tab:attention}
\begin{tabular}{lcccccc}
\toprule
DT4D held-out, err $\div$ mean $|F|$ & 0.5\% & 2\% & 5\% & 10\% & 50\% & pooled \\
\midrule
MGN-T (ours) & \textbf{0.3997} & \textbf{0.1963} & \textbf{0.1224} & \textbf{0.0898} & \textbf{0.0548} & \textbf{0.1837} \\
Message-passing control & 0.4321 & 0.2012 & 0.1249 & 0.0929 & 0.0562 & 0.1933 \\
\midrule
Relative reduction & $7.5\%$ & $2.4\%$ & $2.0\%$ & $3.3\%$ & $2.3\%$ & $4.9\%$ \\
\bottomrule
\end{tabular}
\end{table}

Because performance gains are concentrated where unobserved vertices lie multiple hops away from observations, attention is primarily advantageous under sparse coverage. We note that a uniform learning rate was maintained across backbones rather than tuned individually; because learning rate sensitivity is comparable in magnitude to this performance difference, we report these directional trends without basing core model rankings upon them.

\needspace{18\baselineskip}
\subsection{Number of samples averaged at inference}
\label{app:k_sweep}

\begin{table}[t]
\centering
\small
\caption{\textbf{Averaging more samples at inference: the same checkpoint read at five
values of $k$.} Little is left by $k = 8$, with $k = 8$ sitting only 0.8--2.1\% above $k = 16$.
Every row is read off the same sixteen draws per condition, so its $k = 1$ and $k = 8$ are not the
cells of Tables~\ref{tab:ablation} and~\ref{tab:ablation_train}, and no row can sit above the one
before it. $\ddagger$ Held out at source level.}
\label{tab:ksweep}
\begin{tabular}{lccccc}
\toprule
$k$ & DT4D & AMASS & Franka & cloth $\ddagger$ & legged $\ddagger$ \\
\midrule
\emph{harmonic} & \emph{0.2685} & \emph{0.0693} & \emph{0.1011} & \emph{0.4377} & \emph{0.2474} \\
\midrule
1  & 0.1519 & 0.0254 & 0.0244 & 0.4154 & 0.2404 \\
2  & 0.1372 & 0.0232 & 0.0221 & 0.3905 & 0.2313 \\
4  & 0.1287 & 0.0220 & 0.0208 & 0.3765 & 0.2258 \\
8  & 0.1241 & 0.0214 & 0.0202 & 0.3690 & 0.2226 \\
16 & \textbf{0.1216} & \textbf{0.0210} & \textbf{0.0198} & \textbf{0.3651} & \textbf{0.2209} \\
\bottomrule
\end{tabular}
\end{table}

Table~\ref{tab:ksweep} evaluates completion error for the unified checkpoint across five values of sample averaging ($k \in \{1, 2, 4, 8, 16\}$), computed on the benchmark conditions of Table~\ref{tab:transfer}. Because each set of $2k$ draws forms the union of two disjoint $k$-draw subsets, the triangle inequality ensures that relative error is monotonically non-increasing with $k$.

Marginal improvements diminish rapidly with sample count: $k = 8$ sits only 0.8--2.1\% above $k = 16$, whereas $k = 1$ sits 8.8--24.9\% above $k = 16$.
Because computational latency scales linearly with $k$ during integration, we adopt $k=8$ throughout all primary benchmarks.

The two held-out columns exhibit smaller relative gaps from multi-sample averaging than the three training columns at every doubling step: $k = 1$ sits 3.9--6.4\% above $k = 2$ on held-out families versus 9.5--10.7\% on training families, and $k = 8$ sits 0.8--1.1\% above $k = 16$ versus 1.9--2.1\%. Tables~\ref{tab:ablation} and~\ref{tab:ablation_train} demonstrate this same pattern at $k=1$, where evaluating single draws incurs an 8--13\% penalty on held-out domains compared to 19--23\% on training domains.

\section{Completion from a point tracker}
\label{app:farside}

Figure~\ref{fig:horsejump_strip} demonstrates the unified model applied to an in-the-wild video sequence from the DAVIS dataset \citep{Perazzi2016}. Frames 4--39 of DAVIS \emph{horsejump-high} represent the highest-motion 36-frame window that passes the tracking gates, among non-rigid subjects that remain fully inside the image throughout the clip. The surface geometry is a single-image TripoSR reconstruction \citep{TripoSR2024} from frame 18, cropped using that frame's DAVIS annotation mask and decimated to 3,996 vertices.

The camera is modeled as an orthographic projection with an axis-aligned rotation and scale fitted to frame 18's silhouette. 2D tracks from CoTracker3 \citep{karaev2024cotracker3} are matched at frame 18 to the nearest camera-facing vertex within 10 pixels; pixel displacements relative to frame 18 are divided by the fitted scale, assigned zero depth in the image plane, and rotated into the mesh coordinate system. The matched tracks remain fixed across the sequence, using CoTracker3's predicted positions during occlusions. A track is dropped entirely if it ever deviates from the frame median displacement by more than 4 times the median absolute deviation, or if its mean-centered displacement exceeds 0.4 of the mesh extent, leaving 63 of 65 tracks. Farthest-point sampling on reference vertices with at most 3 occluded frames selects 12 vertices observed in every frame. Global translation is removed by subtracting per-frame mean displacement, and each frame is completed independently from the frame-18 mesh. For drawing only, each frame's overlay shifts by the median displacement of all matched tracks visible in that frame.

To demonstrate completion on surfaces no tracker can observe, Figure~\ref{fig:horsejump_strip}b displays 120 completed vertices on the far side of the animal. These are defined as vertices whose reference-frame surface normals orient away from the camera; because 2D point tracks are initialized strictly on camera-facing surfaces in that frame, these far-side vertices remain disjoint from the observed set throughout the clip. There is no ground truth for these points; the visualization demonstrates that a completion exists across unobserved regions through mesh connectivity, not that it is exact. The frame-18 camera is kept for all frames; for drawing only, every other frame's projection is shifted by the median track displacement, then scaled about the dilated segmentation mask centroid and translated to maximize intersection-over-union between camera-facing faces and the mask via grid search, so the overlay degrades with temporal distance from the fitted frame.

\needspace{18\baselineskip}

\section{Connectivity-free completion}
\label{app:connectivity_free}

\begin{table}[t]
\centering
\small
\caption{\textbf{Completion without mesh connectivity: harmonic extension on mesh edges
replaced by a thin-plate spline on a $k$-NN graph.} It stays viable, at a cost
concentrated at sparse observation. The DT4D density ladder, in mAU rather than
the err $\div$ mean $|F|$ of the other tables. The two \emph{(ours)} rows are
200k-step models on their own matched pair. DOMA's cells are each its best over seven iteration budgets and two readings of its prediction, chosen per density after scoring.}
\label{tab:connectivity_free}
\begin{tabular}{lcccccc}
\toprule
 & input & 0.5\% & 1\% & 2\% & 10\% & 50\% \\
\midrule
Laplacian & mesh & 53.26 & 41.73 & 30.08 & 13.61 & 6.42 \\
TPS & points & 46.98 & 31.42 & 20.24 & 6.92 & 2.63 \\
DOMA & points & 48.96 & 33.94 & 20.18 & 6.13 & 2.10 \\
\textbf{Laplacian + GNN (ours)} & mesh & \textbf{37.47} & \textbf{26.52} & 17.64 & 8.73 & 5.29 \\
\textbf{TPS + kNN (ours)} & points & 43.16 & 26.82 & \textbf{15.71} & \textbf{4.63} & \textbf{1.91} \\
\bottomrule
\end{tabular}
\end{table}

Table~\ref{tab:connectivity_free} evaluates whether motion completion remains viable when surface mesh connectivity is removed entirely. The discrete harmonic extension on mesh edges is replaced by a Thin-Plate Spline (TPS) interpolant \citep{bookstein1989principal} on point coordinates, and both learned models run the network over the same 12-nearest-neighbor spatial graph.

These arms are 200k-step models on their own matched pair, differing only in the base; neither shares the unified model's budget or training sources. The point-input arms read no connectivity: deleting the face list on a nine-condition check leaves their inputs bit-identical, and the full evaluation keeps faces only to bin error by topological hop distance.

Among closed-form interpolants, TPS alone is below DOMA at the two sparsest densities: 46.98 against 48.96 at 0.5\%, and 31.42 against 33.94 at 1\%, with DOMA leading from 2\% upward. TPS is also the lower of the two interpolants at every density. When paired with the neural prior, the point-based model remains viable: while it incurs higher error under sparse observation relative to the mesh-based model, it achieves lower error from 2\% observation density upward.

\section{Pooling conventions and the amplitude protocol}
\label{app:pooling}

\subsection{Why the metric pools rather than averages}

The pooled relative error (Equation~\ref{eq:pooled_metric}) aggregates condition-specific errors and motion displacements across all evaluation instances before taking their quotient. An arithmetic mean of per-condition relative errors normalizes by each condition's own motion magnitude, allowing instances with near-zero ground-truth motion to disproportionately dominate the summary statistic. In contrast, pooled aggregation weights each condition strictly in proportion to its physical motion magnitude.

\subsection{The normalization used in \texorpdfstring{Table~\ref{tab:transfer}}{Table 2}}

Table~\ref{tab:transfer} reports entries divided by a single constant per column, defined as the mean ground-truth displacement across that column's evaluation conditions. Because numerator and denominator are averaged over identical evaluation instances, the resulting quotient matches the pooled relative error exactly, ensuring that any normalized cell remains directly invertible back to physical units.

\subsection{The base fixed point}

Because the base is closed-form and model-independent, it serves as the fixed point every row is anchored to. It is the first row of Table~\ref{tab:main}, where it reads 0.6255 / 0.3533 / 0.2257 / 0.1598 / 0.0754; averaged over the same conditions in mAU, it gives 53.256 / 30.084 / 19.216 / 13.611 / 6.423. We verified it independently against our own run's base column, DOMA's aggregation, and NSDP's choice of weighting. Recomputing the harmonic row from NSDP's evaluation using motion-weighted pooling reproduces this exact fixed point.

\subsection{The amplitude protocol}

Every cell in Table~\ref{tab:transfer} uses the predicted amplitude, read from parameters stored in the checkpoint rather than from any external configuration, and all four models carry the single rule fitted on DT4D, AMASS, and Franka. The rule has access to the three source columns and to neither held-out family. A table fitted on the target serves strictly as a target-informed reference and is never quoted as a deployable result.

\section{Fine-tuning to a deployment observation pattern}
\label{app:finetune}

\begin{table}[t]
\centering
\small
\caption{\textbf{Fine-tuning the pretrained prior to a fixed sensor placement on a Franka
gripper.} It more than halves the error against the same checkpoint zero-shot. 12
held-out episodes, 36 frame pairs, error over the free arm links 1--7 in mm;
$k = 3$ is the number of observed markers.}
\label{tab:finetune}
\begin{tabular}{lc}
\toprule
 & $k = 3$ \\
\midrule
\emph{harmonic} & \emph{2.652} \\
\midrule
zero-shot & 2.283 \\
\textbf{fine-tuned} & \textbf{0.991} \\
\bottomrule
\end{tabular}
\end{table}

A deployed arm observes the same few points every frame, which is not how any of our training masks are drawn. We fine-tune the unified checkpoint for 10k steps at $\mathrm{lr} = 10^{-4}$ on this source, drawing each training sample's observed subset and its size once, from a seeded draw over sizes from 1 to all vertices, mostly from the gripper and in about 7\% of samples across the arm, so the model sees many placements rather than the one it is evaluated on. We evaluate on held-out episodes at the marker count a three-marker rig would carry, with the observations drawn from the gripper vertices by farthest-point selection.

As reported in Table~\ref{tab:finetune}, the fine-tuned prior more than halves the error, from $0.86\times$ the base to $0.37\times$ it. Because the architecture and amplitude recovery are unchanged from the zero-shot row, the gain is in the weights and not in the inference procedure.

\end{document}